\documentclass[conference]{IEEEtran}
\IEEEoverridecommandlockouts

\usepackage{cite}
\usepackage{amsmath,amssymb,amsfonts}
\usepackage{algorithm}
\usepackage{algorithmic}
\usepackage{graphicx}
\usepackage{textcomp}
\usepackage{xcolor}
\usepackage{url}
\usepackage{booktabs}
\usepackage{multirow}
\usepackage{subcaption}
\usepackage{placeins}
\usepackage{flushend}

\graphicspath{{figure/}}
\providecommand{\Description}[1]{}

\begin{document}

\title{TriHead-GAN: A Generative Adversarial Network with Triple-Head Discriminator for Carbon Emission Time Series Generation}

\author{\IEEEauthorblockN{Zesen Wang, Lijuan Lan\textsuperscript{*}, Yonggang Li, Chunhua Yang}
\IEEEauthorblockA{\textit{School of Automation, Central South University} \\
Changsha, China \\
lijuan.lan@csu.edu.cn}
\thanks{\textsuperscript{*}Corresponding author.}
\thanks{Source code are available at \protect\url{https://github.com/SanMuGuo/TriHead-GAN}.}
}

\maketitle
\thispagestyle{plain}
\pagestyle{plain}

\begin{abstract}
Accurate carbon emission monitoring is critical for climate policy and emerging regulatory mechanisms such as the EU Carbon Border Adjustment Mechanism, yet city-level high-frequency monitoring data remain extremely scarce, severely limiting data-hungry deep learning models. Time series generation is a natural remedy, but existing GAN and diffusion-based generators often provide limited explicit supervision for the domain structure of carbon emission data: they may match marginal distributional statistics while insufficiently preserving cross-variable correlations between CO$_2$ and co-emitted pollutants and meteorological factors, and tend to collapse the first-difference statistics of atmospheric measurements, producing sequences that are smooth on average but lack the realistic step-wise variability of the underlying signals. We propose TriHead-GAN, a Transformer-based adversarial framework whose triple-head discriminator jointly supervises three complementary aspects of the joint distribution: distributional authenticity via a Wasserstein critic, cross-variable dependency via leakage-free regression of the target variable, and step-wise temporal smoothness via adjacent-difference prediction. The generator combines global self-attention with local temporal convolution, per-step noise injection, and an anti-smoothing loss that matches first-difference statistics. Experiments on the self-collected Changsha Carbon dataset, two public carbon datasets (China, US), and the ETTh1 benchmark show that TriHead-GAN achieves favorable performance over mainstream baselines on the vast majority of settings, and that the resulting synthetic windows improve downstream forecasting accuracy in low-resource carbon monitoring scenarios.
\end{abstract}

\begin{IEEEkeywords}
Data Augmentation, Time Series Generation, Generative Adversarial Network, Triple-Head Discriminator, Carbon Emission 
\end{IEEEkeywords}

\section{Introduction}

Climate change is one of the most pressing challenges facing humanity, largely driven by greenhouse gas emissions, particularly carbon dioxide (CO$_2$). The Paris Agreement aims to hold the global temperature increase well below 2$^\circ$C above pre-industrial levels and to pursue efforts to limit the increase to 1.5$^\circ$C, making carbon emission reduction a central policy priority worldwide~\cite{unfccc2015paris}. Annual scientific assessments such as the Global Carbon Budget~\cite{friedlingstein2023global} and high-resolution emission inventories such as EDGAR~\cite{crippa2020edgar} document the persistent gap between emission targets and observed trends. Accurate monitoring and prediction of carbon emissions are therefore critical for carbon trading markets, regional emission reduction policies, and carbon neutrality pathway planning. Beyond science, carbon monitoring is now also tightly coupled with direct economic consequences: the European Union's Carbon Border Adjustment Mechanism (CBAM) entered its definitive regime on 1 January 2026, under which authorised CBAM declarants must annually declare the embedded emissions of covered imports, with the corresponding certificate obligations phasing in over the subsequent years; when actual emissions cannot be adequately determined, default values are applied, raising compliance costs and therefore making verified, high-quality emissions data economically valuable~\cite{ec2024cbam,eu2023cbamreg}. High-quality, high-resolution monitoring data are thus a prerequisite both for scientific carbon management and for industrial competitiveness, which in turn requires reliable predictive models built on top of such data.

Progress in such predictive models, however, is bottlenecked by the scarcity of training data, which constitutes a concrete applied data-mining problem: city-level monitoring stations rarely accumulate enough high-frequency history to train modern deep learners, yet downstream regulators and policymakers still need reliable predictive pipelines. Recent deep learning methods~\cite{ma2025energy,hong2026stable,jin2024carbon} achieve strong nonlinear modeling but remain highly dependent on data scale and quality, and the publicly available carbon emission corpora (e.g., Carbon Monitor~\cite{liu2020carbon}) are mostly aggregated at the national level and far from sufficient for data-intensive learners. Compared to meteorology or power systems, which have decades of high-frequency observations, urban-scale carbon emission monitoring suffers from high deployment cost, sparse spatial coverage, and limited historical records. The recent emergence of time series foundation models such as Sundial and TimeMoE~\cite{liu2025sundial,shi2024time} further raises the data bar: domain-specific foundation models for carbon emissions typically require millions of training samples, making data scarcity a fundamental obstacle to further progress. Time series generation has therefore become a natural and pragmatic remedy.

GAN-based methods (e.g., TimeGAN~\cite{yoon2019time}, TTS-GAN~\cite{li2022tts}) and diffusion-based methods (e.g., Diffusion-TS~\cite{yuan2024diffusion}) have shown promising performance on general time series generation tasks. However, when applied to carbon emission scenarios, two practical issues remain unresolved. First, cross-variable consistency is lacking: CO$_2$ exhibits strong correlations with co-emitted pollutants (CO, NO$_2$, O$_3$) and meteorological factors (temperature, humidity, pressure), yet existing discriminators mainly assess overall sample authenticity, allowing generated samples to match marginal distributions while potentially distorting the correlation structure. Second, temporal dynamics are easily distorted: atmospheric variables change slowly and continuously, but Transformer-based generators tend to over-smooth via global attention, and the lack of explicit constraints on local variation can also produce unrealistic abrupt changes. Although grounded in carbon emission data, these two requirements are general properties of most multivariate time series: temporal dependency spans both global patterns and local abrupt variations, while cross-variable dependency lies in the internal joint structure among variables rather than their marginal distributions alone. We therefore target both properties directly and validate the framework beyond carbon data on the general-purpose ETTh1 benchmark.

To address the above issues, this paper proposes TriHead-GAN, a Transformer-based adversarial framework with a triple-head discriminator tailored for multivariate carbon emission time series generation. The discriminator routes the input through three parallel task-specific 1D-CNN branches, each feeding its own head: \textbf{D-Head} evaluates distributional authenticity via the Wasserstein objective; \textbf{R-Head} constrains cross-variable dependencies through leakage-free regression of the target variable from non-target variables; and \textbf{T-Head} enforces the temporal smoothness of atmospheric parameters by predicting adjacent-step differences. The generator combines self-attention with local temporal convolution and per-step noise injection, and is further regularized by an anti-smoothing loss that matches the first-difference statistics. We evaluate TriHead-GAN on the self-collected Changsha Carbon dataset, two public national-level datasets (China Carbon, US Carbon), and the ETTh1 benchmark, and observe consistent improvements over competitive baselines.

Our main contributions are:
\begin{itemize}
  \item We construct \emph{TriHead-GAN}, a Transformer-based adversarial framework specifically designed for multivariate carbon emission time series generation under data scarcity.
  \item We propose a \emph{triple-head discriminator} that simultaneously supervises distributional authenticity, cross-variable dependency, and step-wise temporal smoothness, providing cross-variable consistency-aware multi-aspect supervision.
  \item We present a systematic applied study on a self-collected real-world urban carbon monitoring station and three public datasets, evaluating distribution fidelity, diversity, downstream forecasting utility, and deployment cost, and showing favorable performance over six competitive baselines.
\end{itemize}

\section{Related Work}

\subsection{Carbon Emission Forecasting}

Carbon emission forecasting has evolved from statistical and machine learning models to end-to-end deep learning and foundation-model approaches~\cite{jin2024carbon,ma2025energy,hong2026stable,maji2025carbonx}. Despite continuous advances, existing studies predominantly rely on national-level data from platforms such as Carbon Monitor~\cite{liu2020carbon}, while city-level high-frequency monitoring data remain severely scarce. This bottleneck motivates our focus on high-quality time series generation for low-resource carbon monitoring.

\subsection{Time Series Generation}

Existing approaches can be broadly divided into GAN-based methods that learn temporal distributions through adversarial training, and diffusion-based or likelihood-based methods that model data distributions through denoising or latent-variable inference.

\textbf{GAN-based methods.} RCGAN establishes recurrent conditional generation and the TSTR paradigm~\cite{esteban2017real}, TimeGAN adds embedding-space supervision~\cite{yoon2019time}, and TTS-GAN introduces Transformer-based adversarial generation~\cite{li2022tts}. Recent studies also show that GAN-based synthetic time series can improve downstream forecasting in data-scarce regimes and continue to refine sequence-feature modeling~\cite{chatterjee2025gan,hou2025dlgan}.

\textbf{Diffusion-based methods.} TimeVAE learns latent representations via variational autoencoding~\cite{desai2021timevae}, while Diffusion-TS, PAD-TS, and TimeDP use denoising objectives with frequency-domain, population-aware, and domain-prompt designs~\cite{yuan2024diffusion,li2025population,huang2025timedp}; newer extensions further address irregular sampling and graph-structured spectral relations~\cite{fadlon2025diffusion,shen2025tsgdiff}.

Despite strong general performance, existing generators usually emphasise sample authenticity or denoising fidelity, with limited explicit supervision for domain-specific cross-variable structure and step-wise variation. In carbon emission scenarios, synthetic windows must preserve both the relationships between CO$_2$ and meteorological/pollutant variables and the realistic first-difference statistics of atmospheric measurements. This motivates the dedicated discriminator design pursued in this work.

\section{Methodology}

\subsection{Problem Formulation}

Given a set of real carbon emission time series $\mathcal{X} = \{\mathbf{x}^{(i)}\}_{i=1}^{N}$, where each sample $\mathbf{x}^{(i)} \in \mathbb{R}^{T \times F}$ is a window of length $T$ with $F$ feature variables, our goal is to learn a generative model $G$ that produces synthetic windows $\hat{\mathbf{x}} = G(\mathbf{z}) \in \mathbb{R}^{T \times F}$ from random noise $\mathbf{z} \sim \mathcal{N}(0, \mathbf{I})$, satisfying three criteria: (1)~\textit{distribution consistency} (the generated distribution $p_G$ should approximate $p_{\text{data}}$); (2)~\textit{cross-variable consistency} (inter-variable correlations should conform to the underlying domain structure) and (3)~\textit{temporal coherence} (variation rates and smoothness should match real sequences).

\subsection{Overall Framework}

\begin{figure}[!t]
\centering
  \includegraphics[width=\columnwidth]{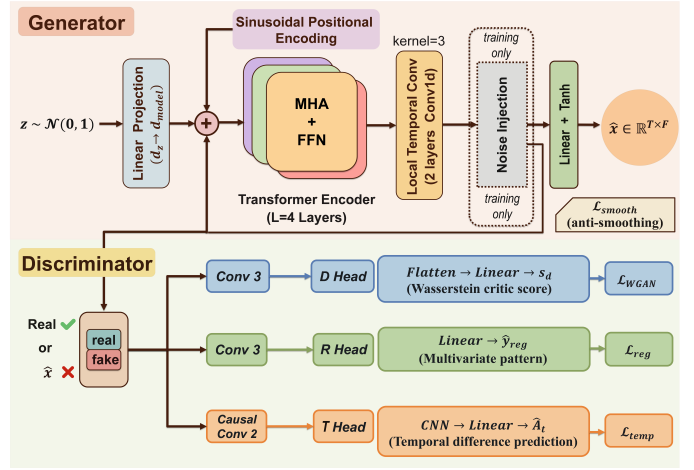}
\Description{Architecture diagram of TriHead-GAN with a Transformer generator and a discriminator containing authenticity, regression, and temporal heads.}
\caption{Overview of TriHead-GAN. The Transformer generator emits time series that the discriminator processes through three parallel CNN branches, each feeding its own head: D-Head (WGAN authenticity), R-Head (cross-variable regression with leakage-free input), and T-Head (causal temporal coherence).}
\label{fig:architecture}
\end{figure}

TriHead-GAN consists of a Transformer-based generator and a Triple-Head Discriminator (Fig.~\ref{fig:architecture}).

\textbf{Transformer-based Generator $G$.} It takes random noise $\mathbf{z} \in \mathbb{R}^{T \times d_z}$ as input, applies linear projection, sinusoidal positional encoding, and a Transformer encoder to extract global temporal dependencies, followed by a Local Temporal Convolution module for fine-grained local dynamics and per-step noise injection for temporal diversity, ultimately generating synthetic segments $\hat{\mathbf{x}} \in \mathbb{R}^{T \times F}$.

\textbf{Triple-Head Discriminator $D$.} It takes time series segments (real or generated) as input and routes them to three task-specific heads, each fed by its own parallel CNN branch. \emph{D-Head} ($D_d$) serves as a Wasserstein critic on top of a dedicated 3-layer 1D-CNN with spectral normalization~\cite{miyato2018spectral}. \emph{R-Head} ($D_r$) has its own 3-layer CNN branch fed only with non-target features, providing leakage-free input for cross-variable regression. \emph{T-Head} ($D_t$) employs a separate 2-layer causal CNN branch to enforce temporal coherence by predicting adjacent-step differences.

\subsection{Transformer-based Generator}

The generator $G$ uses a Transformer encoder as backbone, augmented with local temporal convolution and per-step noise injection.

\textbf{Input projection.} Each time step's noise vector is linearly mapped from the noise dimension $d_z$ to the model dimension $d_{\text{model}}$: $\mathbf{h}_0 = \text{Linear}(\mathbf{z}) \in \mathbb{R}^{T \times d_{\text{model}}}$.

\textbf{Positional encoding.} Sinusoidal positional encoding~\cite{vaswani2017attention} injects absolute temporal information:
\begin{equation}
\begin{aligned}
\text{PE}(t, 2k) &= \sin(t / 10000^{2k/d_{\text{model}}}),\\
\text{PE}(t, 2k{+}1) &= \cos(t / 10000^{2k/d_{\text{model}}}).
\end{aligned}
\end{equation}
The position-injected feature is then $\mathbf{h}_1 = \text{Dropout}(\mathbf{h}_0 + \text{PE})$.

\textbf{Transformer encoding.} The input passes through $L$ Transformer encoder layers, each comprising multi-head self-attention (MHA) and a feed-forward network (FFN), so that $\mathbf{h}_{l+1} = \text{TransformerEncoderLayer}(\mathbf{h}_l)$ for $l = 1, \ldots, L$. The self-attention mechanism enables the generator to perceive information from all time steps, capturing periodic patterns and long-range dependencies in carbon emission data.

\textbf{Local temporal convolution.} The global attention of Transformers tends to over-smooth outputs and lose fine-grained local variation. We add a 2-layer 1D convolution module after the encoder with a residual connection:
\begin{equation}
\begin{aligned}
\mathbf{h}_{\text{local}} &= \text{Conv1d}_2(\text{GELU}(\text{Conv1d}_1(\mathbf{h}_{L+1}))),\\
\mathbf{h}' &= \mathbf{h}_{L+1} + \mathbf{h}_{\text{local}},
\end{aligned}
\end{equation}
where Conv1d uses kernel size 3 with padding to preserve sequence length.

\textbf{Per-step noise injection.} During training, learnable-scale random noise is injected at each time step to enhance temporal diversity:
\begin{equation}
\mathbf{h}'' = \mathbf{h}' + s \cdot \boldsymbol{\epsilon}, \quad \boldsymbol{\epsilon} \sim \mathcal{N}(0, \mathbf{I}), \quad s \in \mathbb{R}^{d_{\text{model}}},
\end{equation}
where $s$ is initialized to $0.05$ and is disabled during inference.

\textbf{Output projection.} Features are mapped back to the data dimension with a Tanh activation that constrains outputs to $[-1, 1]$: $\hat{\mathbf{x}} = \tanh(\text{Linear}(\mathbf{h}'')) \in \mathbb{R}^{T \times F}$.

\subsection{Triple-Head Discriminator}

The Triple-Head Discriminator is the core innovation of TriHead-GAN, evaluating time series quality from three complementary perspectives. Conceptually, the three heads target the marginal, conditional, and transition components of the joint sequence distribution, respectively.

\textbf{Three parallel CNN branches.} The discriminator routes the input $\mathbf{x}$ through three task-specific 1D-CNN branches that operate in parallel, each feeding its own head:
\begin{equation}
\mathbf{h} = \text{CNN}_d(\mathbf{x}), \;\; \mathbf{g} = \text{CNN}_r(\tilde{\mathbf{x}}), \;\; \mathbf{u} = \text{CNN}_\tau(\mathbf{x}),
\end{equation}
where $\text{CNN}_d$ is a 3-layer spectrally-normalized 1D-CNN feeding D-Head, $\text{CNN}_r$ is a separate 3-layer 1D-CNN whose \emph{leakage-free} input $\tilde{\mathbf{x}} = \mathbf{x}_{:, 1:F-1}$ excludes the target column, and $\text{CNN}_\tau$ is a 2-layer causal 1D-CNN feeding T-Head. Each block stacks convolution, LeakyReLU, and LayerNorm. Spectral normalization~\cite{miyato2018spectral} on $\text{CNN}_d$ enforces Lipschitz continuity and complements gradient penalty for stable WGAN training, while the causal padding of $\text{CNN}_\tau$ prevents future information from leaking into the temporal-difference prediction.

\textbf{D-Head: authenticity evaluation.} Following WGAN-GP~\cite{gulrajani2017improved}, D-Head maps its branch features to a scalar critic score: $s_d = D_d(\mathbf{h}) = \text{Linear}(\text{Flatten}(\mathbf{h})) \in \mathbb{R}$.

\textbf{R-Head: structural consistency.} R-Head imposes cross-variable structural regularities. To prevent the target from leaking into its own prediction, $\text{CNN}_r$ receives only the non-target features $\tilde{\mathbf{x}} = \mathbf{x}_{:, 1:F-1}$. R-Head then regresses the target variable at each step:
\begin{equation}
\hat{y}_t^{\text{reg}} = \text{Linear}(\mathbf{g}_t), \;\; \mathcal{L}_{\text{reg}} = \frac{1}{T} \sum_{t=1}^{T} (\hat{y}_t^{\text{reg}} - x_{t,F})^2.
\end{equation}

\textbf{T-Head: temporal coherence.} T-Head consumes the causal-CNN features $\mathbf{u}$ from its own branch to predict adjacent-step differences:
\begin{equation}
\hat{\Delta}_t = D_t(\mathbf{u}_t) \in \mathbb{R}^{F}, \;\; \mathcal{L}_{\text{temp}} = \frac{1}{(T{-}1) F} \sum_{t=1}^{T-1} \| \hat{\Delta}_t - (\mathbf{x}_{t+1} - \mathbf{x}_t) \|^2.
\end{equation}

\subsection{Training Objective}

TriHead-GAN is trained under WGAN-GP with the auxiliary losses introduced above and linearly warmed-up weights.

\textbf{Anti-smoothing loss.} To counter over-smoothed Transformer outputs, we match both the mean and standard deviation of the per-feature absolute first-difference distribution:
\begin{equation}
\begin{aligned}
\mathcal{L}_{\text{smooth}} \;=\; & \,\text{MSE}\!\left( \mu_{\hat{\Delta}}, \, \mu_{\Delta} \right) + \text{MSE}\!\left( \sigma_{\hat{\Delta}}, \, \sigma_{\Delta} \right), \\
\mu_{\Delta} \;=\; & \, \frac{1}{(T{-}1)B} \sum_{b,t} |\mathbf{x}_{t+1}^{(b)} - \mathbf{x}_t^{(b)}|, \\
\sigma_{\Delta}^{2} \;=\; & \, \frac{1}{(T{-}1)B} \sum_{b,t} \bigl(|\mathbf{x}_{t+1}^{(b)} - \mathbf{x}_t^{(b)}| - \mu_{\Delta}\bigr)^{2},
\end{aligned}
\end{equation}
with $\mu_{\hat{\Delta}}$ and $\sigma_{\hat{\Delta}}$ defined analogously on the generated batch. This complements T-Head by matching temporal variation at the statistical level.

\textbf{Discriminator loss:}
\begin{equation}
\begin{aligned}
\mathcal{L}_D =&\; D_d(\hat{\mathbf{x}}) - D_d(\mathbf{x}) + \lambda_{\text{gp}} \, \text{GP} \\
& + \alpha_w \mathcal{L}_{\text{reg}}^{\text{real}} + \beta_w \mathcal{L}_{\text{reg}}^{\text{fake}} + \delta_w (\mathcal{L}_{\text{temp}}^{\text{real}} + \mathcal{L}_{\text{temp}}^{\text{fake}}).
\end{aligned}
\end{equation}

\textbf{Generator loss:}
\begin{equation}
\mathcal{L}_G = -D_d(\hat{\mathbf{x}}) + \gamma_w \mathcal{L}_{\text{reg}}^{\text{fake}} + \delta_w \mathcal{L}_{\text{temp}}^{\text{fake}} + \eta_w \mathcal{L}_{\text{smooth}},
\end{equation}
where warmup weights follow $w_i(e)=w_i^{\ast}\min(1,e/E_w)$ for $w_i \in \{\alpha,\beta,\gamma,\delta,\eta\}$. Each minibatch uses $n_{\text{critic}}$ critic updates, one generator update, and an EMA update on $G$; $G_{\text{EMA}}$ is used at inference.

\subsection{Necessity of the Three Heads}\label{sec:stat_view}

We now show why the three heads are jointly necessary and not redundant, both from a statistical and an empirical-gradient perspective.

\textbf{Statistical view.} The heads supervise three different functionals of the joint window distribution $p(\mathbf{x}_{1:T})$. D-Head approximates the Lipschitz dual of the Wasserstein-1 distance between $p_{\text{data}}$ and $p_G$, sensitive to the \emph{marginal} of full windows but less so to conditional and transition moments. R-Head, predicting the target $x_F$ from non-target features, adds a discrepancy on the \emph{conditional} $p(x_F \mid x_1, \dots, x_{F-1})$, which a generator with the right marginal but distorted cross-variable structure cannot match. T-Head, predicting the one-step difference $x_{t+1}-x_t$, supervises the lag-one \emph{transition kernel} $p(\mathbf{x}_{t+1} \mid \mathbf{x}_{\le t})$ tied to local smoothness; the anti-smoothing loss complements it by matching the first two moments of the absolute first-difference distribution. These three functionals (marginal, conditional, transition) emphasize different aspects of the joint distribution, indicating complementary rather than overlapping supervision.

\textbf{Empirical gradient orthogonality.} To verify that the corresponding training signals are also non-redundant in parameter space, for each of $50$ randomly sampled minibatches on Changsha we compute the generator gradient of the adversarial, regression, and temporal losses and report pairwise cosine similarities: the means are $-0.075_{\pm 0.217}$ (Adv.\ vs.\ Reg.), $0.024_{\pm 0.141}$ (Adv.\ vs.\ Temp.), and $0.066_{\pm 0.133}$ (Reg.\ vs.\ Temp.), all within $\pm 0.08$ of zero with batch-level values bounded by $\pm 0.4$. Cosine similarities clustered near zero are an empirical signature of near-orthogonal gradient directions, indicating that the three heads supply complementary rather than redundant supervisory directions. Combined with the ablation results in Sec.~\ref{sec:ablation}, R-Head and T-Head therefore act as \emph{complementary regularizers} that constrain views of the joint distribution the adversarial loss may not fully capture.

\FloatBarrier
\section{Experiments}

\begin{table}[t]
\caption{Experimental datasets. All use sliding window $T = 24$ with stride $12$.}\label{tab:datasets}
\centering
\setlength{\tabcolsep}{4pt}
\footnotesize
\begin{tabular}{@{}llrcc@{}}
\toprule
Dataset & Source & Time Points & Granularity & Features  \\
\midrule
Changsha & Monitoring station & 19{,}484 & 15\,min & 5 \\
China & Carbon Monitor & 2{,}193 & Daily & 7 \\
US & Carbon Monitor & 2{,}466 & Daily & 7 \\
ETTh1 & Informer benchmark & 17{,}421 & 1\,h & 7 \\
\bottomrule
\end{tabular}
\end{table}

\begin{table*}[!t]
\caption{Main generation quality (lower is better). Per dataset, best and second-best mean values are in bold and underlined; standard deviations over five seeds are reported alongside.}\label{tab:main}
\centering
\setlength{\tabcolsep}{4.5pt}
\footnotesize
\begin{tabular}{@{}ll cc cc cc cc cc@{}}
\toprule
\multirow{2}{*}{Dataset} & \multirow{2}{*}{Method} & \multicolumn{2}{c}{DS$\downarrow$} & \multicolumn{2}{c}{PS$\downarrow$} & \multicolumn{2}{c}{MMD$\downarrow$} & \multicolumn{2}{c}{FID$\downarrow$} & \multicolumn{2}{c}{ACF$\downarrow$} \\
\cmidrule(lr){3-4} \cmidrule(lr){5-6} \cmidrule(lr){7-8} \cmidrule(lr){9-10} \cmidrule(lr){11-12}
& & Mean & Std & Mean & Std & Mean & Std & Mean & Std & Mean & Std \\
\midrule
\multirow{7}{*}{Changsha} & TimeGAN & $0.357$ & $0.076$ & $0.098$ & $0.013$ & $0.047$ & $0.012$ & $2.43$ & $0.34$ & $0.106$ & $0.030$ \\
 & RCGAN & $0.309$ & $0.063$ & $0.042$ & $0.008$ & $\underline{0.028}$ & $0.016$ & $0.83$ & $0.30$ & $0.080$ & $0.016$ \\
 & TTS-GAN & $0.261$ & $0.025$ & $0.032$ & $0.004$ & $0.099$ & $0.021$ & $2.75$ & $0.70$ & $0.033$ & $0.004$ \\
 & Diffusion-TS & $0.336$ & $0.039$ & $0.027$ & $0.001$ & $0.057$ & $0.025$ & $1.20$ & $0.55$ & $0.061$ & $0.012$ \\
 & PAD-TS & $0.296$ & $0.072$ & $\underline{0.026}$ & $0.001$ & $0.048$ & $0.029$ & $1.07$ & $0.56$ & $0.040$ & $0.009$ \\
 & TimeDP & $\underline{0.205}$ & $0.039$ & $\mathbf{0.025}$ & $0.001$ & $0.033$ & $0.018$ & $\underline{0.74}$ & $0.36$ & $\mathbf{0.027}$ & $0.003$ \\
 & \textbf{TriHead-GAN} & $\mathbf{0.0093}$ & $0.013$ & $0.029$ & $0.002$ & $\mathbf{0.00091}$ & $0.0004$ & $\mathbf{0.09}$ & $0.04$ & $\underline{0.031}$ & $0.008$ \\
\midrule
\multirow{7}{*}{China} & TimeGAN & $0.137$ & $0.029$ & $0.126$ & $0.009$ & $0.077$ & $0.116$ & $3.69$ & $1.68$ & $0.080$ & $0.018$ \\
 & RCGAN & $0.169$ & $0.077$ & $0.119$ & $0.007$ & $\underline{0.029}$ & $0.016$ & $1.82$ & $0.55$ & $\underline{0.054}$ & $0.006$ \\
 & TTS-GAN & $0.245$ & $0.110$ & $0.125$ & $0.009$ & $0.067$ & $0.064$ & $2.99$ & $1.38$ & $0.094$ & $0.011$ \\
 & Diffusion-TS & $0.186$ & $0.091$ & $0.115$ & $0.008$ & $0.055$ & $0.028$ & $2.08$ & $0.51$ & $0.096$ & $0.007$ \\
 & PAD-TS & $\underline{0.130}$ & $0.072$ & $\underline{0.106}$ & $0.013$ & $0.039$ & $0.024$ & $\underline{1.57}$ & $0.48$ & $0.064$ & $0.007$ \\
 & TimeDP & $0.214$ & $0.057$ & $0.116$ & $0.010$ & $0.042$ & $0.026$ & $1.58$ & $0.42$ & $0.110$ & $0.007$ \\
 & \textbf{TriHead-GAN} & $\mathbf{0.045}$ & $0.024$ & $\mathbf{0.101}$ & $0.005$ & $\mathbf{0.0029}$ & $0.001$ & $\mathbf{0.37}$ & $0.06$ & $\mathbf{0.048}$ & $0.007$ \\
\midrule
\multirow{7}{*}{US} & TimeGAN & $\underline{0.090}$ & $0.072$ & $0.108$ & $0.005$ & $0.035$ & $0.017$ & $3.15$ & $0.84$ & $0.156$ & $0.019$ \\
 & RCGAN & $0.269$ & $0.033$ & $0.114$ & $0.012$ & $0.045$ & $0.018$ & $2.43$ & $0.68$ & $0.111$ & $0.007$ \\
 & TTS-GAN & $0.174$ & $0.064$ & $0.107$ & $0.007$ & $0.041$ & $0.015$ & $2.03$ & $0.67$ & $\underline{0.100}$ & $0.017$ \\
 & Diffusion-TS & $0.212$ & $0.052$ & $0.111$ & $0.015$ & $0.070$ & $0.027$ & $2.63$ & $0.56$ & $0.114$ & $0.005$ \\
 & PAD-TS & $0.176$ & $0.047$ & $\underline{0.102}$ & $0.009$ & $0.047$ & $0.020$ & $1.89$ & $0.60$ & $\mathbf{0.082}$ & $0.009$ \\
 & TimeDP & $0.203$ & $0.094$ & $0.106$ & $0.005$ & $\underline{0.030}$ & $0.016$ & $\underline{1.65}$ & $0.31$ & $0.122$ & $0.001$ \\
 & \textbf{TriHead-GAN} & $\mathbf{0.0064}$ & $0.006$ & $\mathbf{0.092}$ & $0.004$ & $\mathbf{0.0027}$ & $0.001$ & $\mathbf{0.49}$ & $0.02$ & $\mathbf{0.082}$ & $0.003$ \\
\midrule
\multirow{7}{*}{ETTh1} & TimeGAN & $0.327$ & $0.167$ & $0.087$ & $0.012$ & $0.032$ & $0.007$ & $2.24$ & $0.28$ & $0.098$ & $0.031$ \\
 & RCGAN & $0.395$ & $0.054$ & $0.058$ & $0.009$ & $\underline{0.031}$ & $0.010$ & $1.10$ & $0.11$ & $0.034$ & $0.011$ \\
 & TTS-GAN & $0.293$ & $0.070$ & $0.055$ & $0.010$ & $0.065$ & $0.037$ & $1.22$ & $0.47$ & $0.046$ & $0.012$ \\
 & Diffusion-TS & $0.262$ & $0.075$ & $0.044$ & $0.008$ & $0.060$ & $0.049$ & $1.04$ & $0.78$ & $0.052$ & $0.009$ \\
 & PAD-TS & $0.271$ & $0.124$ & $\underline{0.041}$ & $0.004$ & $0.031$ & $0.011$ & $0.74$ & $0.18$ & $0.036$ & $0.005$ \\
 & TimeDP & $\underline{0.210}$ & $0.044$ & $0.043$ & $0.005$ & $0.032$ & $0.015$ & $\underline{0.63}$ & $0.23$ & $\underline{0.031}$ & $0.006$ \\
 & \textbf{TriHead-GAN} & $\mathbf{0.013}$ & $0.006$ & $\mathbf{0.038}$ & $0.002$ & $\mathbf{0.00070}$ & $0.0001$ & $\mathbf{0.06}$ & $0.005$ & $\mathbf{0.019}$ & $0.001$ \\
\bottomrule
\end{tabular}
\end{table*}

\subsection{Experimental Setup}

\textbf{Datasets.} We evaluate TriHead-GAN on four multivariate datasets summarized in Table~\ref{tab:datasets}: our self-collected Changsha Carbon urban monitoring data, two national-level carbon emission datasets (China Carbon and US Carbon from Carbon Monitor~\cite{liu2020carbon}), and the public ETTh1 benchmark. The Changsha Carbon dataset is collected from a real-world urban atmospheric monitoring station from late June 2025 to mid-January 2026 at a 15-minute sampling cadence, with stable sensor operation and negligible missing values across the recording period. It reflects a common low-resource setting in city-level carbon monitoring: high-frequency CO$_2$ observations are expensive to deploy and maintain, and the available history is much shorter than that of mature meteorological benchmarks. It contains $12$ raw variables: PM2.5, PM10, CO, NO$_2$, SO$_2$, O$_3$, CO$_2$, wind speed, wind direction, temperature, humidity, and atmospheric pressure. With CO$_2$ as the target variable, we retain the $5$ variables whose Pearson correlation with CO$_2$ exceeds $|r| = 0.3$, namely CO, PM2.5, PM10, wind speed, and CO$_2$. The remaining datasets contain $7$ variables. In every dataset the R-Head target is the canonical last column, CO$_2$ for the three carbon datasets and oil temperature (OT) for ETTh1, following each source's standard target convention. All datasets use sliding windows of length $T = 24$ and stride $12$, and preprocessing includes missing-value interpolation, outlier repair, MinMax normalization to $[0, 1]$, and a linear mapping to $[-1, 1]$ during training to match the generator output.

\textbf{Baselines.} We compare against six representative time series generators, organised into two families. \emph{GAN-based methods}: TimeGAN~\cite{yoon2019time}, an embedding-space GAN with a supervised reconstruction loss; RCGAN~\cite{esteban2017real}, a recurrent conditional GAN that established the TSTR evaluation paradigm and TTS-GAN~\cite{li2022tts}, a Transformer-based GAN tailored to time series. \emph{Diffusion-based methods}: Diffusion-TS~\cite{yuan2024diffusion}, a denoising diffusion model with a Fourier-domain loss; PAD-TS~\cite{li2025population}, a population-aware diffusion model that captures subpopulation heterogeneity and TimeDP~\cite{huang2025timedp}, a multi-domain diffusion model with domain prompts.

\textbf{Metrics.} Generation quality is measured by five complementary metrics, all in lower-is-better form. \emph{Discriminative score} (DS) is the deviation from $0.5$ of a post-hoc real/fake classifier's error rate; values near $0$ mean real and synthetic windows are indistinguishable. \emph{Predictive score} (PS) is the MAE of a one-step-ahead forecaster trained on synthetic and tested on real data (predicting the final step of each window from the preceding steps), probing whether the conditional structure needed for downstream prediction is preserved. \emph{Maximum mean discrepancy} (MMD) is the kernel distance between real and generated samples, sensitive to higher-order moments. \emph{Fr\'echet distance} (FID) is the Wasserstein-$2$ distance between Gaussian fits to real and generated features, summarising overall distribution alignment. \emph{Autocorrelation difference} (ACF) is the absolute gap between real and generated autocorrelation curves over multiple lags, reflecting temporal-dependence fidelity. All five metrics are computed by a fixed external protocol applied identically to every method and independent of the TriHead-GAN discriminator: DS trains a separate two-layer LSTM classifier and PS a two-layer GRU forecaster, while MMD and FID are computed in the raw flattened-window feature space (FID being the Fr\'echet distance between Gaussian fits, without any learned embedding).

\textbf{Implementation details.} The generator uses $d_z = 64$, $d_{\text{model}} = 128$, $8$ attention heads, $4$ Transformer encoder layers, and feed-forward dimension $256$. The discriminator uses $128$ hidden channels, a 3-layer spectrally-normalized CNN for the D-Head branch, a separate 3-layer CNN for the R-Head branch, and a 2-layer causal CNN for the T-Head branch. We train for $1000$ epochs with batch size $32$, Adam ($\beta_1 = 0.5$, $\beta_2 = 0.9$), learning rate $10^{-4}$ with cosine annealing, $n_{\text{critic}} = 5$, gradient clipping at $1.0$, EMA decay $0.999$, and a 300-epoch auxiliary-loss warmup. Loss weights are $\lambda_{\text{gp}} = 10$, $\alpha = 3$, $\beta = 1$, $\gamma = 3$, $\delta = 1$, and $\eta = 2$. All baselines are paper-aligned reimplementations developed with reference to the original papers and their open-source repositories, where hyperparameters are configured following the recommended values in the original papers. To control variance, the main comparison is repeated over five seeds (42, 43, 44, 45, 46), while all other experiments (ablation, TSTR, diversity, convergence, sensitivity) are repeated over three seeds (42, 43, 44); reported numbers are the seed-averaged values. All models are trained on a single NVIDIA RTX 4060Ti GPU.

\subsection{Main Results}

As shown in Table~\ref{tab:main}, TriHead-GAN attains the best (or tied-best) score in $18$ of the $20$ (dataset, metric) cells and is uniformly first on MMD, FID, and DS across all four datasets. We attribute this advantage to two design choices: the triple-head discriminator spreads adversarial pressure across distribution, cross-variable, and temporal aspects, helping the generator balance multiple statistics more effectively, and the anti-smoothing loss penalises high-frequency collapse, keeping short-range power and autocorrelation aligned with the real signal.

On the joint-distribution metrics MMD and FID, TriHead-GAN improves clearly over the best baseline on every dataset, and it ties with PAD-TS for the best ACF on US. The two cells in which TriHead-GAN is not first are Changsha PS ($0.029$ vs.\ TimeDP $0.025$) and Changsha ACF ($0.031$ vs.\ TimeDP $0.027$); both are univariate marginal statistics on which the strongest diffusion baselines (TimeDP, PAD-TS) are competitive. These gains stem from the proposed design rather than from any single tuned statistic: the regression and temporal heads add explicit cross-variable and step-wise supervision on top of the Wasserstein critic, and the anti-smoothing loss preserves the first-difference statistics, so the generator matches the joint structure that marginal-oriented baselines model less directly.

The two model families display complementary strengths. Diffusion- and likelihood-based generators are trained with a denoising objective that directly fits fine-grained per-variable distributional detail, which makes them naturally strong on the univariate marginal statistics: configured with the recommended settings from their original papers and open-source code, TimeDP attains the best PS and ACF on Changsha and the lowest DS among baselines on Changsha and ETTh1, and PAD-TS ties our model for the best ACF on US. Adversarial training instead optimises a sample-level discrimination signal that is better suited to capturing the joint structure of the data, which TriHead-GAN reinforces with explicit cross-variable and step-wise supervision through its R- and T-heads. Consequently, TriHead-GAN keeps the lead on the joint-distribution metrics, ranking first on DS, MMD, and FID on every dataset, while remaining comparable to the strongest diffusion baselines on the marginal statistics.

\subsection{Significance Analysis}\label{sec:significance}

\begin{table}[tb]
\caption{Number of metrics (out of $5$) on which \textbf{TriHead-GAN} is significantly lower than each baseline ($p<0.05$, paired one-sided $t$-test).}\label{tab:significance_matrix}
\centering
\setlength{\tabcolsep}{3.4pt}
\footnotesize
\begin{tabular}{@{}lcccccc@{}}
\toprule
\multirow{2}{*}{Dataset} & \multicolumn{6}{c}{TriHead-GAN vs.\ baseline (sig.\ / 5)} \\
\cmidrule(lr){2-7}
 & TimeGAN & RCGAN & TTS-GAN & Diff.-TS & PAD-TS & TimeDP \\
\midrule
Changsha & 5/5 & 5/5 & 3/5 & 4/5 & 3/5 & 3/5 \\
China & 4/5 & 4/5 & 5/5 & 5/5 & 4/5 & 5/5 \\
US & 5/5 & 5/5 & 4/5 & 5/5 & 3/5 & 5/5 \\
ETTh1 & 5/5 & 5/5 & 5/5 & 4/5 & 5/5 & 4/5 \\
\midrule
Total & 19/20 & 19/20 & 17/20 & 18/20 & 15/20 & 17/20 \\
\bottomrule
\end{tabular}
\end{table}

To confirm that the gains in Table~\ref{tab:main} are not driven by seed variance, we run a paired one-sided $t$-test for every (dataset, baseline, metric) cell on the same five-seed runs (lower-is-better, $\Delta=\overline{\text{TriHead-GAN}}-\overline{\text{baseline}}$, significant when $\Delta<0$ and $p<0.05$). Table~\ref{tab:significance_matrix} summarises the significant counts per (dataset, baseline) pair.

Across all $4 \times 6 \times 5 = 120$ comparisons, TriHead-GAN is significantly better at $p<0.05$ in $105$ cells. The $15$ non-significant cases are confined to the univariate marginal metrics, eight on PS and six on ACF, plus a single MMD case (China vs.\ TimeGAN), and arise mainly against the strongest baselines (TTS-GAN and the diffusion models PAD-TS, TimeDP, Diffusion-TS), where the absolute gaps on these marginal statistics are small and TriHead-GAN is statistically comparable rather than worse. This split is by design rather than a deficiency: the triple-head discriminator deliberately distributes supervision across the marginal, conditional, and transition components of the joint distribution instead of over-optimising any single univariate marginal, so diffusion baselines tuned toward marginal denoising fidelity are expected to stay competitive on the per-variable statistics (PS, ACF), whereas TriHead-GAN should, and does, dominate the metrics that reflect joint and cross-variable structure. On the joint-distribution metrics the advantage is uniform: TriHead-GAN is significantly better on DS and FID in all $24$ comparisons and on MMD in $23$ of $24$. The improvements reported in Table~\ref{tab:main} are therefore consistent with significant gains on the large majority of metric--baseline pairs. These are paired one-sided tests over five seeds without multiple-comparison correction, so we read them as evidence that the improvements are stable across seeds rather than as strong stand-alone statistical claims.

\subsection{Ablation Study}\label{sec:ablation}

To verify the contribution of each component, we design four ablation variants: (1)~\emph{w/o T}: removing the temporal coherence head ($\delta = 0$); (2)~\emph{w/o R}: removing the regression consistency head ($\alpha = \beta = \gamma = 0$); (3)~\emph{w/o AS}: removing the anti-smoothing loss ($\eta = 0$); (4)~\emph{MLP gen.}: replacing the Transformer$+$LocalConv generator with an MLP.

Table~\ref{tab:ablation} reveals four findings. \emph{(i) R-Head is the most stable structural contributor:} removing it worsens all five metrics on Changsha and China and most on US and ETTh1, with MMD rising $49\%$/$31\%$/$83\%$ on Changsha/China/ETTh1, confirming the value of explicit cross-variable supervision. \emph{(ii) The anti-smoothing loss is an important temporal regularizer:} removing it inflates Changsha MMD by $107\%$ and FID by $125\%$ and degrades ACF/FID on the other datasets, matching its role of preventing local-variation collapse. \emph{(iii) T-Head gives selective gains}, mainly on PS and ACF (removing it degrades Changsha ACF by $39\%$, Changsha PS by $21\%$, and China FID by $39\%$), while DS is occasionally lower without it, as the critic then faces fewer auxiliary constraints. \emph{(iv) The Transformer generator helps more as variables grow:} an MLP wins three Changsha cells ($F = 5$) and stays competitive on a few ACF cells of the $7$-variable datasets, but loses most cells overall, with MMD increasing by $11\%$--$170\%$. The full TriHead-GAN is thus the most balanced configuration, and the per-component degradation matches the gradient orthogonality in Sec.~\ref{sec:stat_view}: each head covers a distinct aspect, so each ablation hurts a recognisable subset of metrics.

\begin{table}[tb]
\caption{Ablation study (lower is better). Best and second-best per metric in bold and underlined. w/o T / w/o R / w/o AS drop the temporal, regression, and anti-smoothing components; MLP gen.\ replaces the Transformer generator with an MLP.}\label{tab:ablation}
\centering
\setlength{\tabcolsep}{2.8pt}
\footnotesize
\begin{tabular}{@{}llccccc@{}}
\toprule
Dataset & Variant & DS$\downarrow$ & PS$\downarrow$ & MMD$\downarrow$ & FID$\downarrow$ & ACF$\downarrow$ \\
\midrule
\multirow{5}{*}{Changsha} & \textbf{Full} & $0.013$ & $\underline{0.028}$ & $\mathbf{0.00087}$ & $\underline{0.08}$ & $\underline{0.028}$ \\
 & w/o T & $\underline{0.0055}$ & $0.034$ & $\underline{0.0011}$ & $0.13$ & $0.039$ \\
 & w/o R & $0.018$ & $0.033$ & $0.0013$ & $0.13$ & $0.038$ \\
 & w/o AS & $0.015$ & $0.038$ & $0.0018$ & $0.18$ & $0.035$ \\
 & MLP gen. & $\mathbf{0.0048}$ & $\mathbf{0.025}$ & $0.0016$ & $\mathbf{0.07}$ & $\mathbf{0.015}$ \\
\midrule
\multirow{5}{*}{China} & \textbf{Full} & $\underline{0.028}$ & $\underline{0.102}$ & $\underline{0.0036}$ & $\underline{0.41}$ & $\underline{0.050}$ \\
 & w/o T & $\mathbf{0.010}$ & $0.104$ & $0.0039$ & $0.57$ & $0.062$ \\
 & w/o R & $0.037$ & $0.109$ & $0.0047$ & $0.65$ & $0.090$ \\
 & w/o AS & $0.039$ & $\mathbf{0.096}$ & $\mathbf{0.0024}$ & $0.56$ & $0.076$ \\
 & MLP gen. & $0.036$ & $0.107$ & $0.0040$ & $\mathbf{0.38}$ & $\mathbf{0.049}$ \\
\midrule
\multirow{5}{*}{US} & \textbf{Full} & $\mathbf{0.0041}$ & $0.093$ & $0.0032$ & $\mathbf{0.50}$ & $\underline{0.082}$ \\
 & w/o T & $0.0098$ & $\mathbf{0.089}$ & $\mathbf{0.0020}$ & $\underline{0.51}$ & $0.091$ \\
 & w/o R & $\underline{0.0088}$ & $0.097$ & $\underline{0.0024}$ & $0.53$ & $0.093$ \\
 & w/o AS & $0.018$ & $\underline{0.091}$ & $0.0025$ & $0.54$ & $0.091$ \\
 & MLP gen. & $0.019$ & $0.094$ & $0.0062$ & $0.55$ & $\mathbf{0.049}$ \\
\midrule
\multirow{5}{*}{ETTh1} & \textbf{Full} & $0.011$ & $\mathbf{0.037}$ & $\mathbf{0.00071}$ & $\mathbf{0.06}$ & $\underline{0.018}$ \\
 & w/o T & $\mathbf{0.0027}$ & $\underline{0.038}$ & $\underline{0.00088}$ & $0.08$ & $0.024$ \\
 & w/o R & $0.014$ & $\underline{0.038}$ & $0.0013$ & $0.08$ & $0.020$ \\
 & w/o AS & $\underline{0.0027}$ & $\mathbf{0.037}$ & $0.00092$ & $\underline{0.07}$ & $0.020$ \\
 & MLP gen. & $0.023$ & $0.039$ & $0.0019$ & $0.12$ & $\mathbf{0.017}$ \\
\bottomrule
\end{tabular}
\end{table}

\subsection{Downstream Prediction Enhancement}

To validate the practical utility of generated data beyond distributional similarity, we conduct Train-on-Synthetic, Test-on-Real (TSTR) experiments~\cite{esteban2017real}, an evaluation paradigm subsequently adopted and extended by PATE-GAN~\cite{jordon2019pategan} and the sample-level audit framework of Alaa et al.~\cite{alaa2022faithful}. The Real+Syn protocol additionally simulates the practical use case where a limited real monitoring history is augmented with synthetic windows before training a downstream forecaster. We evaluate three settings: (1)~\emph{TRTR} (train and test on real data); (2)~\emph{TSTR} (train on synthetic data, test on real); and (3)~\emph{Real+Syn} (train on a size-matched 1:1 mixture, test on real). For Real+Syn, the total number of training windows is kept the same as TRTR, with real and synthetic windows sampled at a 1:1 ratio. We use LSTM, GRU, and Transformer predictors, each predicting the last $6$ steps from the first $18$ steps of a window. This multi-step horizon is more demanding than the one-step predictive score (PS) in Table~\ref{tab:main}, so the two evaluations need not rank methods identically. Table~\ref{tab:tstr} reports MAE averaged over the three predictors.

\begin{table}[tb]
\caption{Downstream prediction utility measured by Train-on-Synthetic-Test-on-Real (TSTR). Values are MAE averaged over LSTM, GRU, and Transformer predictors.}\label{tab:tstr}
\centering
\setlength{\tabcolsep}{4pt}
\footnotesize
\begin{tabular}{@{}llccc@{}}
\toprule
Dataset & Method & TRTR$\downarrow$ & TSTR$\downarrow$ & Real+Syn$\downarrow$ \\
\midrule
\multirow{7}{*}{Changsha} & \textbf{TriHead-GAN} & $0.0470$ & $\mathbf{0.0486}$ & $\underline{0.0419}$ \\
 & TTS-GAN & $0.0470$ & $0.0848$ & $0.0479$ \\
 & PAD-TS & $0.0470$ & $0.0518$ & $0.0420$ \\
 & RCGAN & $0.0470$ & $0.0702$ & $0.0475$ \\
 & TimeGAN & $0.0470$ & $0.1390$ & $0.0488$ \\
 & TimeDP & $0.0470$ & $\underline{0.0502}$ & $\mathbf{0.0411}$ \\
 & Diffusion-TS & $0.0470$ & $0.0523$ & $0.0432$ \\
\midrule
\multirow{7}{*}{China} & \textbf{TriHead-GAN} & $0.0576$ & $\mathbf{0.0483}$ & $\underline{0.0523}$ \\
 & TTS-GAN & $0.0576$ & $0.0696$ & $0.0572$ \\
 & PAD-TS & $0.0576$ & $0.0591$ & $\mathbf{0.0516}$ \\
 & RCGAN & $0.0576$ & $0.0739$ & $0.0556$ \\
 & TimeGAN & $0.0576$ & $0.0996$ & $0.0546$ \\
 & TimeDP & $0.0576$ & $0.0589$ & $0.0565$ \\
 & Diffusion-TS & $0.0576$ & $\underline{0.0546}$ & $0.0524$ \\
\midrule
\multirow{7}{*}{US} & \textbf{TriHead-GAN} & $0.0671$ & $\mathbf{0.0674}$ & $\underline{0.0643}$ \\
 & TTS-GAN & $0.0671$ & $0.0931$ & $0.0705$ \\
 & PAD-TS & $0.0671$ & $\underline{0.0734}$ & $\mathbf{0.0626}$ \\
 & RCGAN & $0.0671$ & $0.0816$ & $0.0665$ \\
 & TimeGAN & $0.0671$ & $0.0991$ & $0.0670$ \\
 & TimeDP & $0.0671$ & $0.0749$ & $0.0645$ \\
 & Diffusion-TS & $0.0671$ & $0.0807$ & $0.0656$ \\
\midrule
\multirow{7}{*}{ETTh1} & \textbf{TriHead-GAN} & $0.0491$ & $\mathbf{0.0435}$ & $\mathbf{0.0422}$ \\
 & TTS-GAN & $0.0491$ & $0.0683$ & $0.0507$ \\
 & PAD-TS & $0.0491$ & $0.0514$ & $0.0466$ \\
 & RCGAN & $0.0491$ & $0.0791$ & $0.0514$ \\
 & TimeGAN & $0.0491$ & $0.1041$ & $0.0503$ \\
 & TimeDP & $0.0491$ & $\underline{0.0504}$ & $\underline{0.0459}$ \\
 & Diffusion-TS & $0.0491$ & $0.0551$ & $0.0470$ \\
\bottomrule
\end{tabular}
\end{table}

TriHead-GAN obtains the best TSTR MAE on all four datasets, and the best Real+Syn MAE on ETTh1; on Changsha, China, and US the best Real+Syn score goes to TimeDP (Changsha) and PAD-TS (China, US), with TriHead-GAN a close second. Notably, under the size-controlled real--synthetic training of the Real+Syn protocol (where the total number of windows matches TRTR), replacing half of the real windows with TriHead-GAN samples still pushes the MAE \emph{below} the TRTR baseline on every dataset, with relative reductions of $11\%$ on Changsha, $9\%$ on China, $4\%$ on US, and $14\%$ on ETTh1, indicating that the synthetic windows carry useful complementary signal.

Across all four datasets, the TSTR MAE of TriHead-GAN improves on the strongest baseline by $3$--$14\%$ ($3\%$ on Changsha, $12\%$ on China, $8\%$ on US, and $14\%$ on ETTh1), suggesting that its cross-variable supervision yields synthetic data that transfers better to the real forecasting task. The strongest baselines stay competitive rather than degenerate (Diffusion-TS and PAD-TS track the TRTR reference closely, e.g., Diffusion-TS reaches $0.0523$ on Changsha and $0.0546$ on China), so the downstream gains are measured against genuinely strong synthetic data.

\subsection{Diversity and Mode Coverage}\label{sec:diversity}

\begin{figure}[tb]
\centering
\includegraphics[width=\columnwidth]{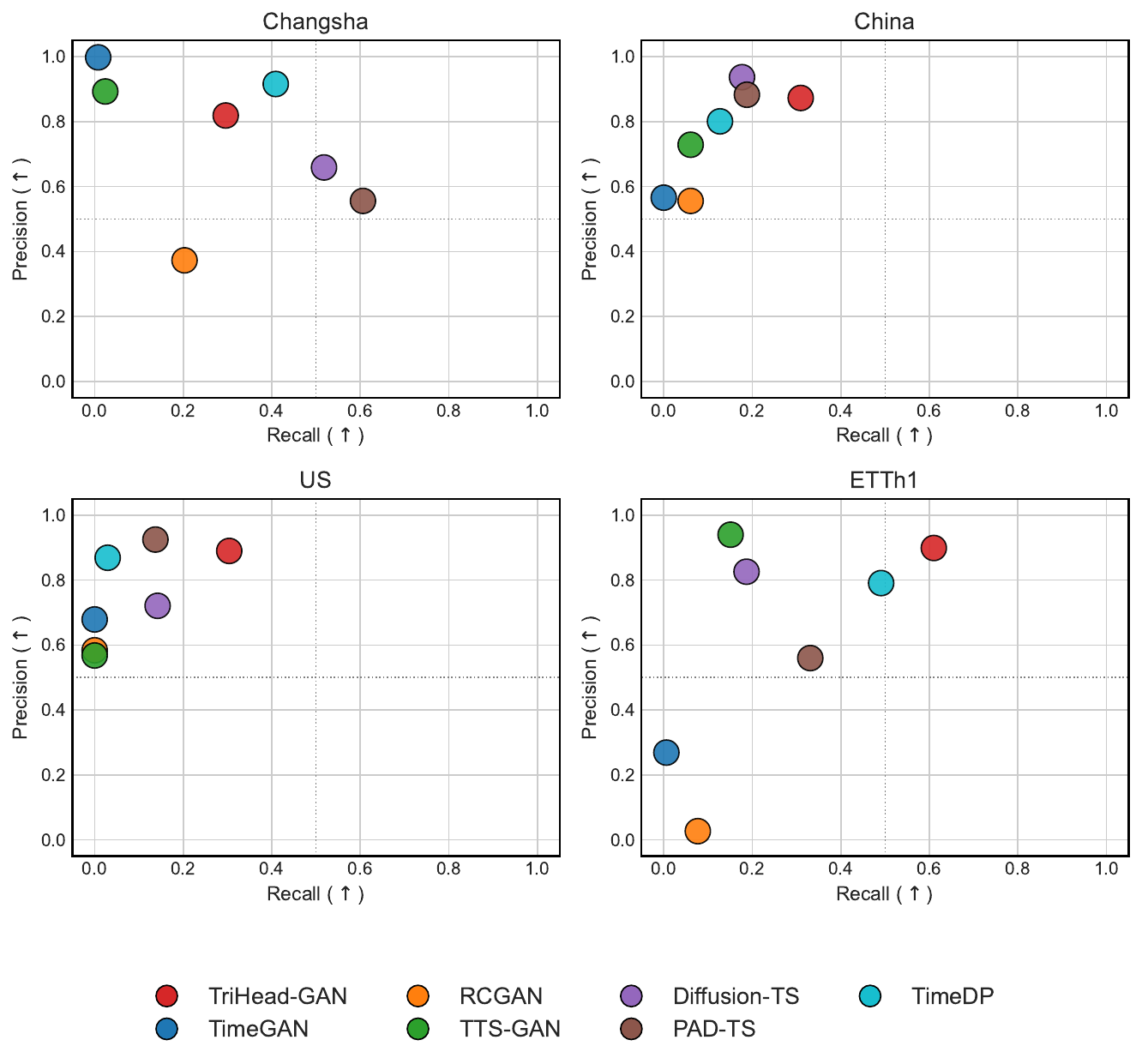}
\Description{Precision-Recall scatter for diversity analysis across four datasets.}
\caption{Diversity analysis on four datasets: Precision (sample fidelity, $y$-axis) vs.\ Recall (mode coverage, $x$-axis).}
\label{fig:diversity}
\end{figure}

Fig.~\ref{fig:diversity} reports the Precision--Recall trade-off using the manifold-based estimator of Kynk{\"a}{\"a}nniemi et al.~\cite{kynkaanniemi2019improved}, refined by the density/coverage criterion of Naeem et al.~\cite{naeem2020reliable}. TriHead-GAN occupies a high-Precision operating point on China, US, and ETTh1 while retaining competitive Recall; on Changsha it sits at a balanced high-Precision, moderate-Recall position (Precision $\approx 0.82$, Recall $\approx 0.30$), where TimeGAN and TTS-GAN reach higher Precision only by collapsing Recall to near zero. Across methods, the Precision-oriented regime targeted by TriHead-GAN remains a practically useful operating point for carbon monitoring, where synthetic windows that stay close to observed trajectories yield more reliable training data for downstream forecasters.

\subsection{Qualitative Visualization}

\begin{figure}[tb]
\centering
\includegraphics[width=\columnwidth]{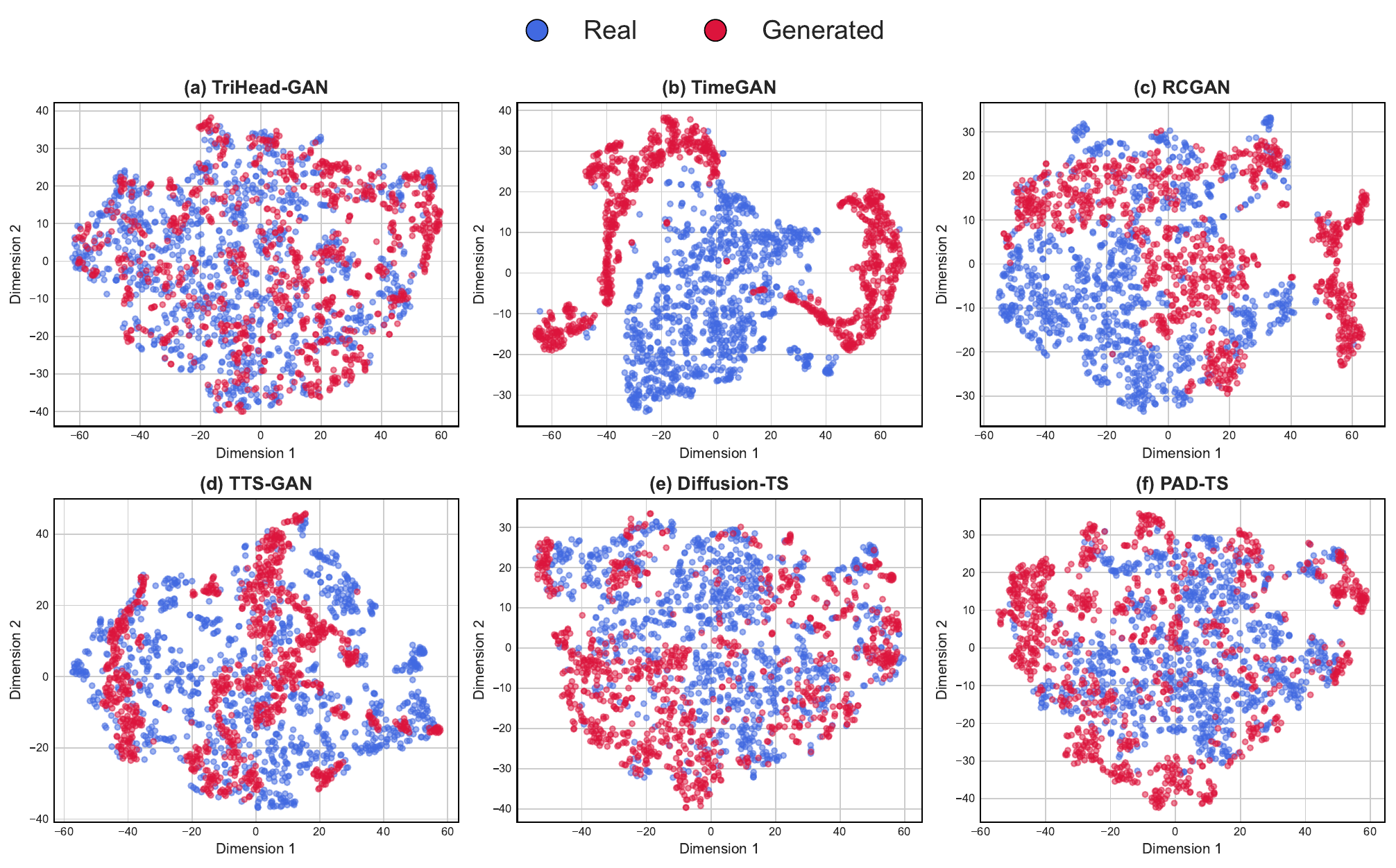}
\Description{t-SNE projections of real vs.\ generated samples on Changsha.}
\caption{t-SNE projections of real (blue) and generated (red) samples on Changsha for TriHead-GAN and five representative baselines.}
\label{fig:tsne}
\end{figure}

\begin{figure}[tb]
\centering
\includegraphics[width=\columnwidth]{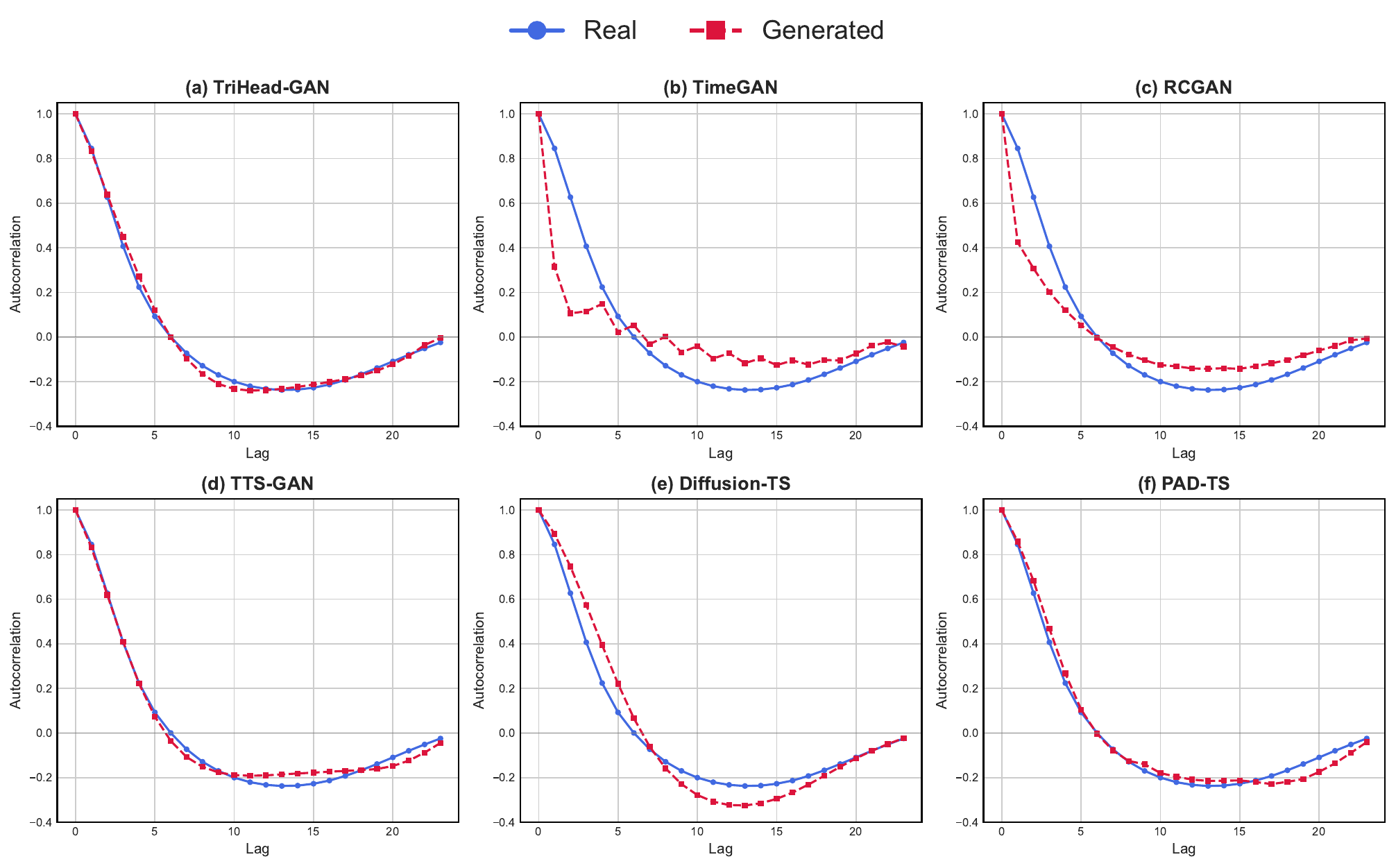}
\Description{Autocorrelation curves of real vs.\ generated sequences on Changsha.}
\caption{Autocorrelation function of real vs.\ generated sequences on Changsha for TriHead-GAN and five representative baselines.}
\label{fig:acf}
\end{figure}

Figs.~\ref{fig:tsne} and \ref{fig:acf} provide qualitative support on Changsha; the same patterns hold on the other three datasets and are omitted for space. The t-SNE projection in Fig.~\ref{fig:tsne} shows that TriHead-GAN samples overlap densely with the real distribution and are scattered across the same multi-mode area as the real points, providing evidence against severe mode collapse. TimeGAN and RCGAN partially separate from the real manifold, while Diffusion-TS spreads beyond it, the spatial counterpart of its near-zero autocorrelation after lag~2 in Fig.~\ref{fig:acf}. The ACF curves further show that TriHead-GAN, TTS-GAN, and PAD-TS are tightly grouped at short lags but diverge at lags above $10$, where the anti-smoothing loss in TriHead-GAN closes the residual gap to the real curve.

\subsection{Convergence and Training Dynamics}\label{sec:convergence}

\begin{figure}[tb]
\centering
\includegraphics[width=\columnwidth]{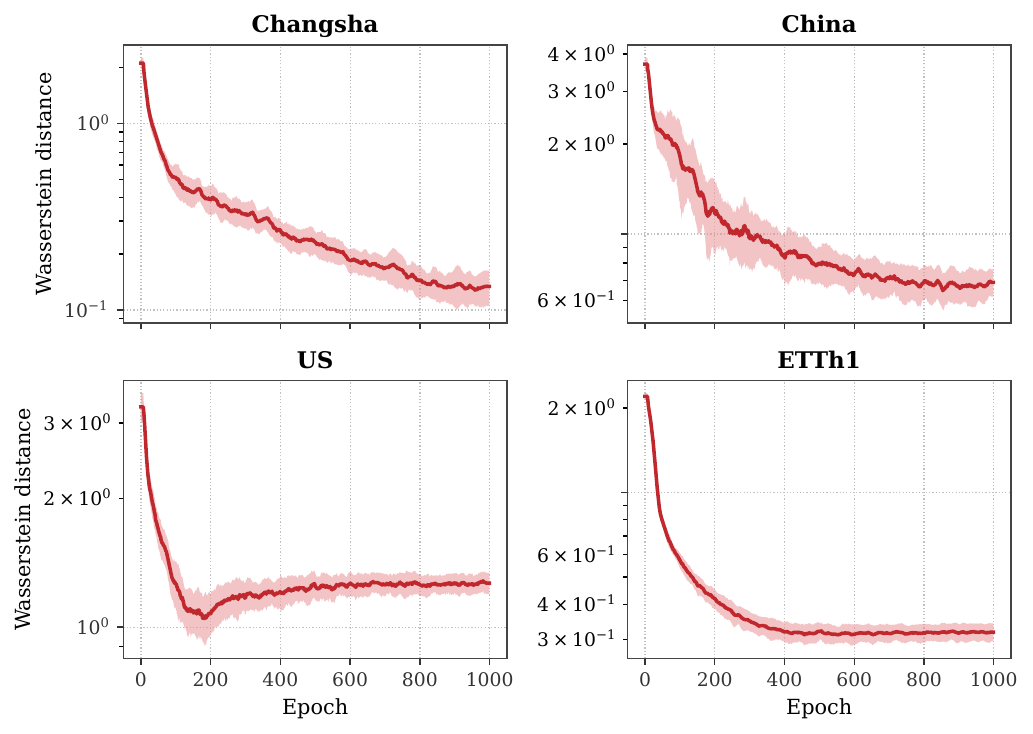}
\Description{Wasserstein distance over training epochs for four datasets.}
\caption{Wasserstein distance estimate during training ($1000$ epochs) on the four datasets.}
\label{fig:convergence}
\end{figure}

Fig.~\ref{fig:convergence} reports the per-epoch Wasserstein-distance estimate from the critic. The relative drop from initial to final values is $96.8\%$ on Changsha, $86.2\%$ on China, $63.9\%$ on US, and $88.9\%$ on ETTh1. The curves show overall decreasing trends with narrow seed bands and little oscillation typical of unstable WGAN training, supporting the design choice of combining gradient penalty with spectral normalization. The descent has a clear two-phase shape: a steep drop over the first $\sim$200 epochs followed by a long plateau, which coincides with the $300$-epoch auxiliary-loss warmup; by the time R-Head and T-Head reach full strength the critic has already absorbed most of the distance, and the auxiliary heads refine rather than dominate the trajectory.

\subsection{Computational Efficiency}\label{sec:efficiency}

\begin{figure}[tb]
\centering
\includegraphics[width=\columnwidth]{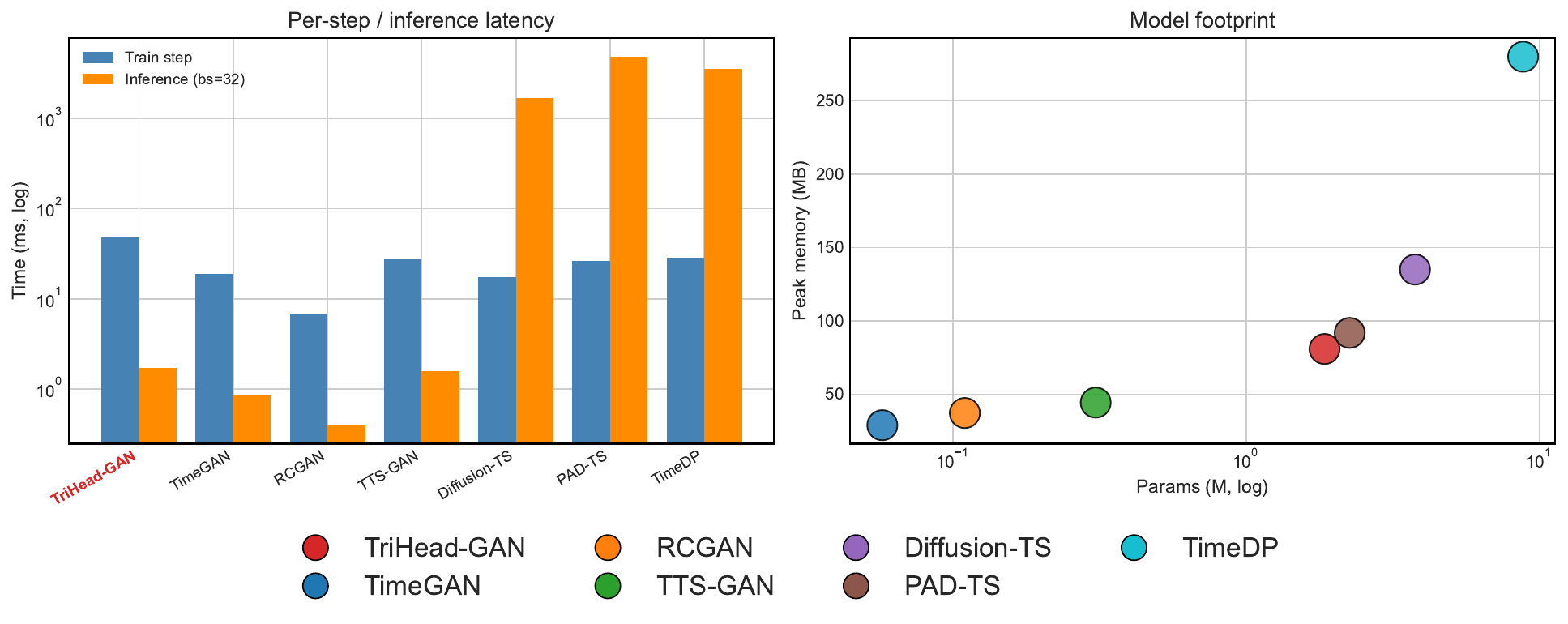}
\Description{Bar chart of training and inference latency, plus scatter of parameter count vs peak memory for all baselines.}
\caption{Computational profile on Changsha. Left: per-step training latency vs.\ batch-32 inference latency (log scale). Right: parameter count (log) vs.\ peak GPU memory.}
\label{fig:efficiency}
\end{figure}

Fig.~\ref{fig:efficiency} summarises the computational profile on Changsha. TriHead-GAN has the largest per-step training time ($47.7$\,ms) among all methods, reflecting the cost of three discriminator branches together with a Transformer generator, though at $1.85$\,M parameters it is smaller than the diffusion baselines (PAD-TS $2.25$\,M, Diffusion-TS $3.76$\,M, TimeDP $8.80$\,M). Its decisive advantage is at inference: sampling a batch of $32$ takes only $1.70$\,ms, three orders of magnitude faster than the diffusion baselines, which require many denoising steps ($1{,}705.7$\,ms for Diffusion-TS, $4{,}866.3$\,ms for PAD-TS, and $3{,}535.1$\,ms for TimeDP). This single-pass sampling makes TriHead-GAN well suited to city-level carbon-monitoring deployments: the generator can be trained offline on station history and repeatedly queried to provide auxiliary synthetic windows for downstream forecasters without replacing real sensor observations, so that the added training cost is paid once while the per-query deployment cost remains on par with single-pass GANs and far below iterative diffusion baselines.

\subsection{Scalability and Sensitivity}\label{sec:scalability}

\begin{figure}[tb]
\centering
\includegraphics[width=\columnwidth]{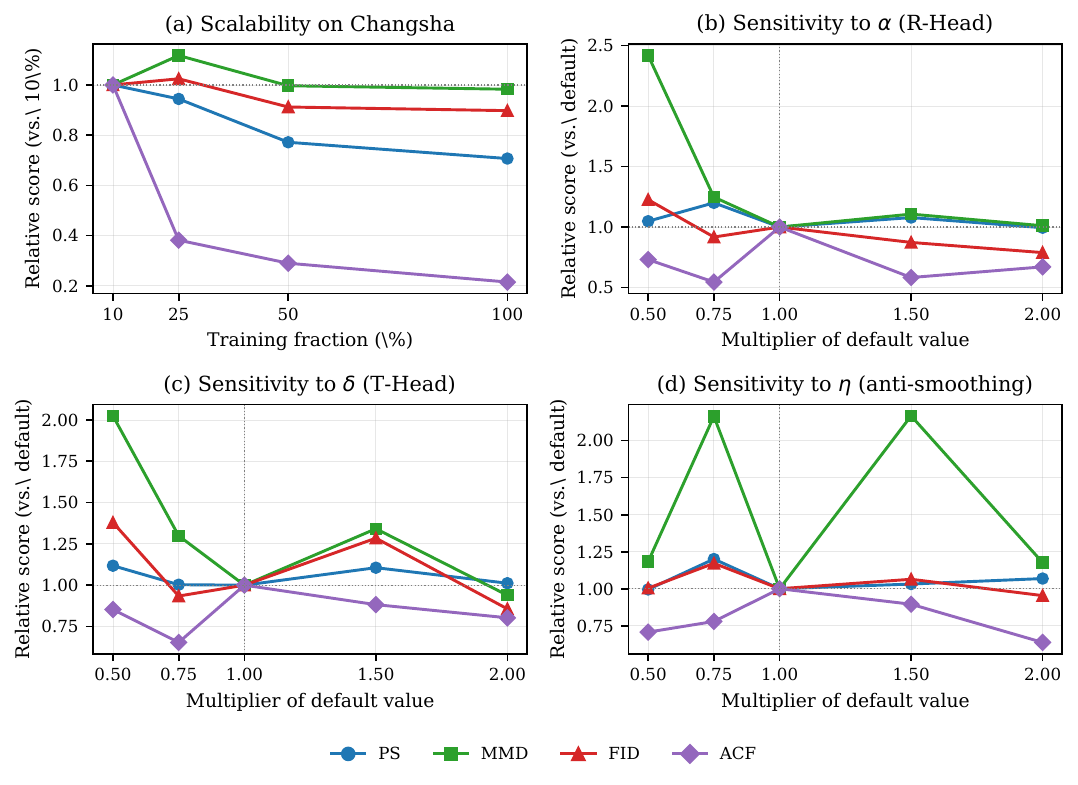}
\Description{Four-panel 2x2 plot showing scalability across training fractions and sensitivity sweeps over the three auxiliary loss weights, with one shared legend.}
\caption{Scalability and loss-weight sensitivity on Changsha (lower is better). (a)~training-fraction sweep $\{10,25,50,100\}\%$ normalised to $10\%$; (b)--(d)~single-weight sweeps over $\alpha$, $\delta$, $\eta$, each normalised to its $1.0\times$ default (dotted line). Each colour denotes one metric.}
\label{fig:sens}
\end{figure}

Panel~(a) of Fig.~\ref{fig:sens} reports the scalability behaviour: PS, FID, and ACF improve monotonically as the training fraction increases, while MMD is noisy below the $25\%$ point and stabilises afterwards. This is consistent with the cross-variable head requiring enough samples to reliably estimate the conditional structure of CO$_2$ given the surrounding meteorological and pollutant variables.

Panels~(b)--(d) sweep the three non-trivial auxiliary weights ($\alpha$ for R-Head, $\delta$ for T-Head, and $\eta$ for anti-smoothing) in $\{0.5, 0.75, 1.0, 1.5, 2.0\}\times$ the defaults (the other two regression weights $\beta,\gamma$ are scaled by the same factor as $\alpha$ when $\alpha$ is swept, since all three weight the regression head, and $\lambda_{\text{gp}}$ uses the standard WGAN-GP value). Three observations follow. (i)~\emph{Stability:} across the full sweep no configuration collapses, with PS, FID, and ACF confined to $[0.028, 0.034]$, $[0.087, 0.152]$, and $[0.021, 0.039]$. (ii)~\emph{Per-weight profile:} larger $\alpha$ keeps improving FID up to $2.0\times$ ($0.135 \to 0.087$); $\delta$ is the most sensitive (its $0.5\times$ setting gives the worst FID of the sweep, $0.152$, so under-weighting transition supervision is the riskiest setting), while $\eta$ is the most robust. (iii)~\emph{Defaults:} the $1.0\times$ setting lies on the Pareto frontier of the four metrics but is not best on every one, since the three heads optimise different functionals.

\FloatBarrier
\section{Conclusion and Limitations}

We propose TriHead-GAN, a Transformer-based adversarial framework for carbon emission time series generation. Its triple-head discriminator simultaneously supervises distributional authenticity, cross-variable dependency, and step-wise temporal smoothness, and the generator is further regularized by local temporal convolution, per-step noise injection, and an anti-smoothing loss. Across four datasets, TriHead-GAN achieves the strongest distribution fidelity (ranking first on DS, MMD, and FID on every dataset) and competitive-to-best downstream forecasting utility, with statistically significant improvements over six representative baselines on the large majority of metric--baseline pairs.

\textbf{Limitations and future work.} TriHead-GAN is not uniformly best: Changsha PS/ACF and US ACF remain competitive but not first, and on China and US it is narrowly second to PAD-TS on the Real+Syn downstream protocol. Two operational failure modes follow from the design: (i)~the cross-variable head needs enough samples to estimate the conditional structure, so TriHead-GAN suits stations with on the order of a thousand or more sliding windows; (ii)~R-Head presupposes a designated target variable, which is natural for carbon monitoring (CO$_2$) but less natural when variables are exchangeable, in which case R-Head can be replaced by a randomly masked target rotation. Future work will pursue this masked rotation and explore integration with time series foundation models for carbon-specific pretraining.

\section*{Acknowledgment}
This work was supported in part by the National Natural Science Foundation of China under Grant 62394341, Grant 62027811. The authors used ChatGPT to assist with drafting and polishing the manuscript text; the proposed methods were produced and verified by the authors.

\bibliographystyle{IEEEtran}
\bibliography{ref}

\end{document}